\PassOptionsToPackage{unicode}{hyperref}
\PassOptionsToPackage{hyphens}{url}
\documentclass[
]{article}
\usepackage{pdflscape}
\usepackage[top=2.5cm,bottom=2.5cm,left=3.5cm,right=3.5cm]{geometry}
\usepackage{xcolor}
\usepackage{amsmath,amssymb}
\usepackage{iftex}
\ifPDFTeX
  \usepackage[T1]{fontenc}
  \usepackage[utf8]{inputenc}
  \usepackage{textcomp} 
\else 
  \usepackage{unicode-math} 
  \defaultfontfeatures{Scale=MatchLowercase}
  \defaultfontfeatures[\rmfamily]{Ligatures=TeX,Scale=1}
\fi
\usepackage{lmodern}
\ifPDFTeX\else
\fi
\IfFileExists{upquote.sty}{\usepackage{upquote}}{}
\IfFileExists{microtype.sty}{
  \usepackage[]{microtype}
  \UseMicrotypeSet[protrusion]{basicmath} 
}{}
\makeatletter
\@ifundefined{KOMAClassName}{
  \IfFileExists{parskip.sty}{%
    \usepackage{parskip}
  }{
    \setlength{\parindent}{0pt}
    \setlength{\parskip}{6pt plus 2pt minus 1pt}}
}{
  \KOMAoptions{parskip=half}}
\makeatother
\usepackage{longtable,booktabs,array}
\usepackage{calc} 
\usepackage{etoolbox}
\makeatletter
\patchcmd\longtable{\par}{\if@noskipsec\mbox{}\fi\par}{}{}
\makeatother
\IfFileExists{footnotehyper.sty}{\usepackage{footnotehyper}}{\usepackage{footnote}}
\makesavenoteenv{longtable}
\usepackage{graphicx}
\makeatletter
\newsavebox\pandoc@box
\newcommand*\pandocbounded[1]{
  \sbox\pandoc@box{#1}%
  \Gscale@div\@tempa{\textheight}{\dimexpr\ht\pandoc@box+\dp\pandoc@box\relax}%
  \Gscale@div\@tempb{\linewidth}{\wd\pandoc@box}%
  \ifdim\@tempb\p@<\@tempa\p@\let\@tempa\@tempb\fi
  \ifdim\@tempa\p@<\p@\scalebox{\@tempa}{\usebox\pandoc@box}%
  \else\usebox{\pandoc@box}%
  \fi%
}
\def\fps@figure{htbp}
\makeatother
\usepackage{bookmark}
\IfFileExists{xurl.sty}{\usepackage{xurl}}{} 
\hypersetup{
  hidelinks,
  pdfcreator={LaTeX via pandoc}}

\title{\LARGE\bfseries Generative Responsible AI Data Evaluation Schema (GRAIDES) for AI Assurance in Local Government}
\author{Ethan Knights$^1$, Christopher Conlan$^1$, Temilorun Gbolahan$^1$, Stephen Waterman$^1$,\\\\ Gurpreet Muctor$^1$\\\\
\small$^1$\textit{AI Innovation Lab, Westminster City Council}}
\date{}

\begin{document}
\maketitle

\section{Abstract}\label{abstract}

Trust in the application of generative Artificial Intelligence (AI)
relies on well-governed measurable evidence of performance \& safety. In
practice, however, evaluation data is often fragmented across systems,
inconsistently structured and difficult to compare. We introduce the
Generative Responsible AI Data Evaluation Schema (GRAIDES) as a
lightweight open-source data model for centralising AI observability
across popular vendors. Practical blueprints for code, architecture and
statistical evaluation are shared as guidance about how to approach
generative system assurance at the organisational level. Illustrative
case study results are reported from Westminster City Council's AI
catalogue with a focus on measuring human-model alignment including
detecting systematic disagreement between evaluators. By framing
evaluations as a data modelling problem, GRAIDES provides a practical
pathway toward more consistent and reproducible benchmarking, tuning and
assurance activities for generative AI systems.

\section{Introduction}\label{introduction}

Generative Artificial Intelligence (AI) has moved rapidly from research
prototypes into mainstream public sector deployment, creating an urgent
need for robust governance that matches the pace of innovation. In
particular, Large Language Models (LLMs) have been widely applied across
education, healthcare and government (e.g., Kasneci et al., 2023;
Thirunavukarasu et al., 2023; Fu et al., 2025). This owes to the
generalised performance of foundation models across summarisation,
classification, conversational and multimodal reasoning tasks (Budnikov
et al., 2025).

The procurement of generative AI across (and even within) public
organisations is fragmented (Ada Lovelace Institute, 2024), exacerbating
governance risks. Without clear and unified guidance, individual teams
may pilot different models for specific use-cases, record metrics
inconsistently or lack the infrastructure to compare performance across
systems and deployments. This piecemeal approach makes it difficult to
answer fundamental questions across scaling use-cases like which models
have been used, under what configurations and to what effect. The
fragmentation produces risks including the undermining of operational
efficiency across diverse domains (e.g., social housing, customer
contact, urban planning), fostering reproducibility crises (e.g., Nosek
et al., 2022) and driving non-adherence to ethical frameworks'
Responsible AI principles (e.g., assuring that AI systems are fair,
inclusive and transparent; Microsoft, 2022). The inconsistency erodes
public trust, as citizens cannot be assured that services are being
delivered fairly, transparently or effectively across different
departments and organisations.

Observability of LLM interactions is a prerequisite to much of the
academic guidance around generative AI adoption (e.g., White et al.,
2024; Lin, 2024; Tian et al., 2025; for overview see Papagiannidis et
al. 2025). To provide assurance that a system meets a given required
standard will always require evaluating model in- and outputs in some
form (e.g., analysing probabilities from systems deployed in production
or success rates across golden dataset batteries). Practically, there
are many engineering recommendations for handling telemetry (e.g., Menon
et al., 2024), language (e.g., Iusztin \& Labonne, 2024; Pahune \&
Akhtar, 2025), vector (e.g., Joshi, 2025) and evaluation data (e.g.,
Saini et al., 2025). Likewise, some attention has been paid to human
factors where, for example, Chen et al. (2025) emphasised benefits of
balancing developer control, transparency and ease of use when building
data models by trading computational efficiency gains for less mental
model complexity.

Here, we introduce the Generative Responsible AI Data Evaluation Schema
(GRAIDES), a vendor‑agnostic, data‑centric framework for organising and
evaluating generative AI activity in operational settings. The GRAIDES
is a relational schema that links system interactions, model and prompt
configurations, and evaluation artefacts into a single auditable
structure. This abstraction is designed to address a recurring practical
challenge in deployed environments: tools, vendors, models, prompts and
evaluators change rapidly in an organisation, but assurance activities
need durable data relationships (e.g., as defined by the GRAIDES) to
enable consistent analyses (e.g., longitudinal cross‑system
comparisons).

Alongside the schema, we present guidance around evaluation pathways, to
illustrate how the framework can be used to support organisational
assurance at different levels of maturity. These range from rapid human
approval workflows to more detailed evaluations (i.e., for unstructured
text and classification model outputs). In all cases, we focus on
enabling an organisation to measure, at scale, the degree of alignment
between human and automated evaluations. We report applied case studies
drawn from Westminster City Council's AI catalogue, showing how GRAIDES
supports both lightweight oversight and deeper diagnostic analysis by
making human and automated evaluations directly comparable. Together,
these contributions position GRAIDES as a practical centralised
foundation for reproducible and auditable evaluations of generative AI
systems in real organisational contexts, like local government.

\section{Schema \& Deployment
Blueprints}\label{schema-deployment-blueprints}

Put simply, the GRAIDES is a data model (Figure 1) designed to link
together data entities that are relevant to generative AI applications.
These entities are reflected within tables (represented here in
capitals): the data model links the source generative \emph{SYSTEMS}
(e.g., a conversational system) with \emph{CHANNELS} (e.g., through web
or telephony) and each channel has independent \emph{CONVERSATIONS} that
can contain many \emph{MESSAGES} (e.g., the message sender can be from a
user, AI or tool). Crucially, these \emph{CONVERSATIONS} and
\emph{MESSAGES} are then enrichable with two types of metadata: model
configurations (describing the source system) and evaluations of
performance/safety (from humans, LLMs or both). Those two types of
metadata are stored in three connected tables. Model configuration
metadata uses: \emph{LLMS}, \emph{LLM\_CONFIGS} \&
\emph{PROMPT\_TEMPLATES.} Evaluations metadata uses: \emph{EVALUATIONS,
EVALUATOR\_METADATA} \& \emph{SCORE\_CONFIGS}.

We balance the schema complexity with its comprehensiveness, to optimise
for covering many scenarios with intuitive Structured Query Language
(SQL) queries that roll important business intelligence indicators up to
the level of, say, a system, channel, LLM or prompt template. This
allows answering simple questions like ``what is the overall system
performance for a given evaluation metric from human grading on a given
month''. Of course, this can be extended for deeper analytics that
isolate system changes (e.g., comparing human or automated grades before
and after deploying a new LLM and prompt template).

To summarise the general process, the GRAIDES relies on a one-off step
to provision specific tables (Figure 1) in a relational SQL database.
Then, preliminary metadata must be registered into the database's
metadata tables (e.g., via an optional application UI) that describe the
entities being used in the organisation (e.g., define the AI systems,
its channels, its LLM and related configuration data like its parameters
and prompt templates). Finally, data-ingestion pipelines can be setup to
routinely convert generative AI source system data into the two key
tables (conversations and messages; see Figure 2A pipeline 1: process)
which should include tagging these with the system metadata registered
above. With those foundations, evaluations data can then be collected
from humans in the business or beyond, as well as from other AI models
(see Figure 2A pipeline 2: evaluate), who all grade the system based on
consistently defined rubrics. In other words, all graders (human or AI)
score the conversations and/or messages on the same defined criteria
(e.g., grade if the model's response is ``approved'' or not) to provide
assurance. See the appendix for more details on the schema's
relationship principles.

\begin{center}
\includegraphics[width=15cm,keepaspectratio]{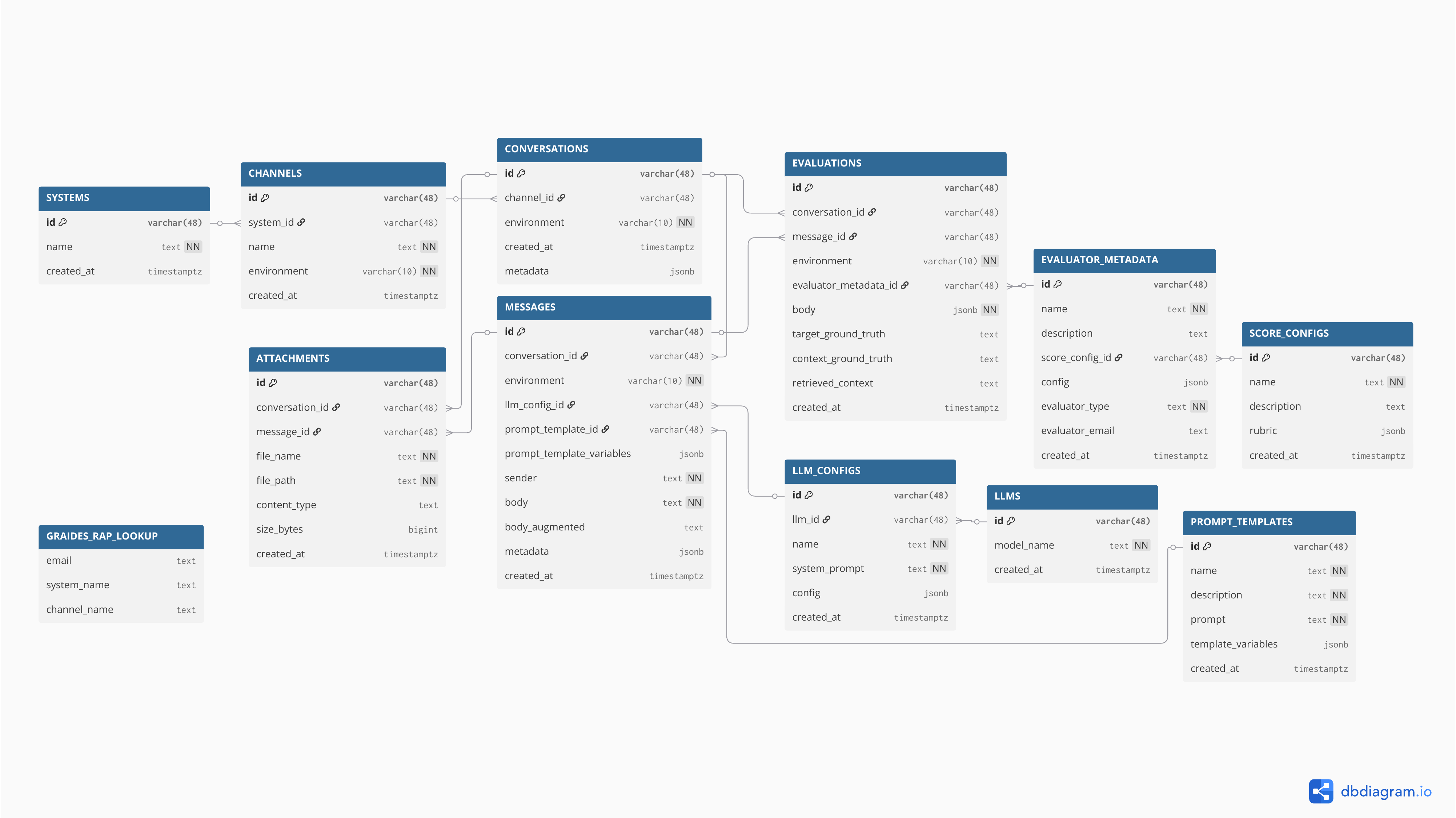}
\end{center}

Figure 1. \textbf{GRAIDES Specification.} Entity Relationship Diagram
(ERD) for a source of truth of an organisation's generative AI activity.

\begin{center}
\includegraphics[width=15cm,keepaspectratio]{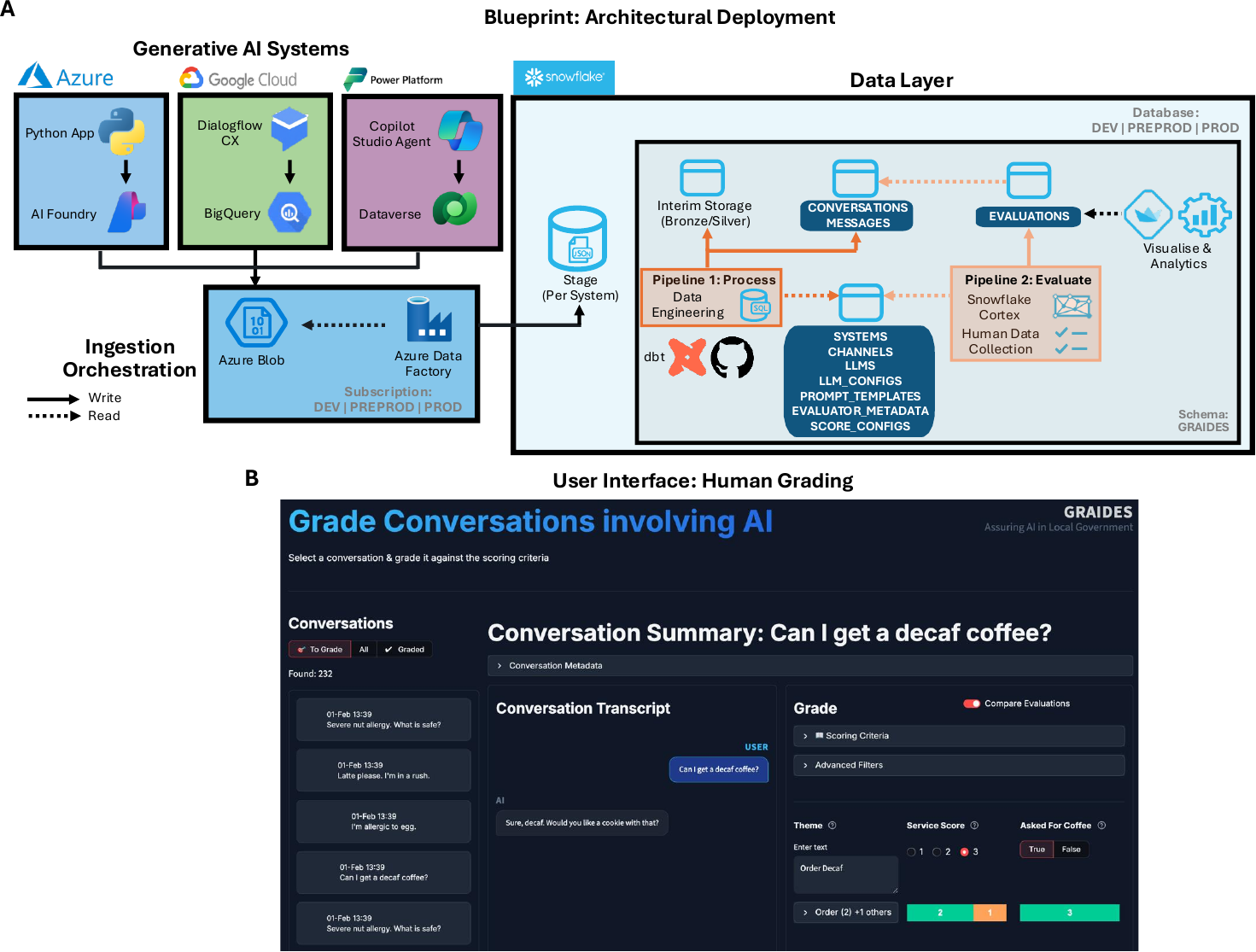}
\end{center}

Figure
2. \textbf{(A}) \textbf{Architectural Blueprint\emph{.}} Data flows from
source AI systems (Azure, Google Cloud, Power Platform) into a central
storage layer (e.g., a managed Snowflake warehouse or a standalone
PostgreSQL container) that hosts GRAIDES components (database tables,
dbt pipelines and a simple Python application). \textbf{(B) Human
Grading UI\emph{.}} A frontend presents conversations and configurable
grading criteria (integer, boolean, or free text) to human graders, with
scoring rubrics; the rubrics are optionally reusable for AI graders.

\section{Evaluations Blueprints}\label{evaluations-blueprints}

The primary incentive of the GRAIDES is to enable systematic management
of assurance activities, by structuring how to consistently log
evaluation data, even when underlying generative AI source systems vary
substantially (e.g., harmonising data from Google Cloud Platform,
Microsoft Azure, Microsoft 365 and custom applications).

As organisational guidance, we present three evaluation blueprints
below: these differ based on (a) output openness (free‑form vs
constrained) and (b) availability of ground truth signals (human labels,
system signals or neither). Common to each pathway, is an ``analytical
layer'' (Figure\,3, bottom) which we highlight because, by governing the
system and evaluation data up front, this layer guarantees better data
readiness for data professionals to perform diagnostic analyses which we
illustrate throughout the Results section (Figures 4-6) in an effort to
show potential return-on-investment from better AI governance.

\subsubsection{Human Approvals (Evaluation Pathway
1)}\label{human-approvals-evaluation-pathway-1}

The first pathway uses a simple human approval workflow which is the
most rapid route to providing some degree of assurance of a system (see
Figure\,3A). In this setup, source system data are first ingested into
the core GRAIDES tables - at a minimum the system/channel, each
conversation and its associated messages.

Human graders can log into the application (or complete a downloadable
evaluation spreadsheet that can be re-uploaded) to grade conversations
using a basic pre‑built ``human approval'' rubric: grade (True/False) if
a conversation is approved. This approach allows organisations to
quickly collect evaluations democratically from a representative group
of staff, providing an early indication of whether a system's outputs
are meeting core expectations.

This approval scoring is intentionally coarse, but it is robust across
different generative use cases. Humans can review, interpret, and flag
problematic outputs regardless of how the system generates content
(e.g., free-text, image/audio, structured classes). This workflow does,
however, become insufficient as adoption scales. As the number of
deployed systems grows, the volume of outputs requiring review outpaces
human oversight. At the same time, growing system complexity often
demands more nuanced evaluation criteria than a binary approval decision
can capture. Therefore, for more comprehensive assurance we next split
the pathways based on the two most common types of generative model
output: unstructured text vs. structured classifications.

\subsubsection{Unstructured Text (Evaluation Pathway
2)}\label{unstructured-text-evaluation-pathway-2}

In the unstructured text pathway, the language model generates free‑form
textual outputs (Figure\,3B, top left). Using the GRAIDES interface
(``admin'' page) users can define an expressive bespoke metric rubric
such as categorising the degree to which model outputs are relevant
(``1: Poor'', ``2: Okay'', ``3: Good'') or if the model exhibited
unnecessary refusal behaviour (``True'', ``False'').

Both human graders and a secondary language‑model evaluator (an
LLM‑Judge) apply this rubric to conversations or individual messages.
Exhaustive human review is rarely feasible at scale, so humans typically
grade only a subset of the data but LLM judges can provide broader
automated coverage. A critical step in the process then, is validating
the automated evaluations by measuring evaluation alignment between
humans and models. This draws on established methods from machine
learning (e.g., examining prediction rates) and the social sciences
(e.g., group comparisons using confidence intervals). The results are
surfaced in visualisations (Figure\,3) and optionally exported for more
rigorous analyses (e.g., do they meet predefined business thresholds?;
how aligned are human/model graders?). Below are four characteristic
human-model alignment patterns that typically emerge (also see
Figure\,3B):

\begin{quote}
Green: High human and high automated scores, indicate an acceptably
performing system that can be reliably evaluated automatically.

Orange: High human but low automated scores, indicate acceptable system
performance but unreliable automated evaluation.

Yellow: Low human and low automated scores, indicate system deficiencies
that are consistently detected.

Red: Low human but high automated scores, indicate problematic
misalignment in which automated evaluations mask underlying system
deficiencies.
\end{quote}

\subsubsection{Structured Classification (Evaluation Pathway
3)}\label{structured-classification-evaluation-pathway-3}

The final pathway concerns systems that produce structured
classifications (Figure\,3, top right). Here, an evaluation rubric must
be created that matches the system's constrained output classes (for
example, classifying inputs into categories A, B, or C) and human
graders simply grade as before. Importantly, we cannot simply apply an
LLM‑Judge approach like earlier (because it simply replicates the
classification task being performed by the source system) so two
evaluation scenarios arise.

In the first evaluation scenario, system‑generated labels or behavioural
signals are available from the source system (e.g., user actions, like a
click that confirms a selected category). These signals can be treated
as evaluation outputs and compared against human grades using the same
alignment framework described earlier. In the second scenario, no
system‑level indicators exist (e.g., as when processing static survey
responses). In this case, assurance relies entirely on human grading,
meaning that the GRAIDES application is useful for collecting human
evaluations. Both scenarios broadly mirror conventional machine‑learning
validation practice, where human-/automated-labelled samples serve as
ground truth for calculating typical classification metrics. Ultimately,
statistical tests can be used to assess if system performance falls
significantly below acceptable levels according to human and/or AI
evaluations (e.g., in particular, an organisation may want to identify
if there are any red failure patterns in Figure\,3, where positive AI
evaluations are masking issues that have been detected by humans).

\begin{center}
\includegraphics[width=15cm,keepaspectratio]{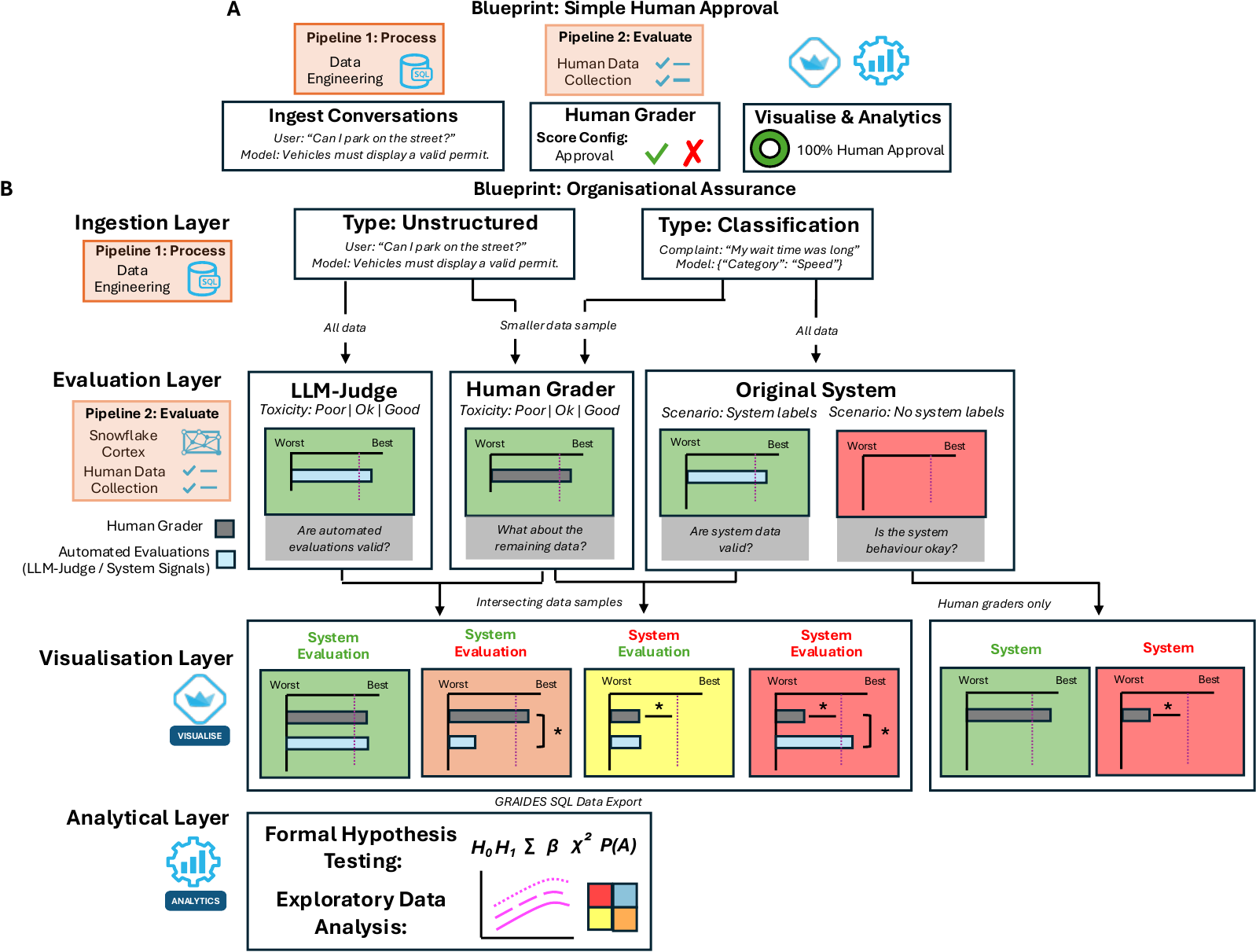}
\end{center}

Figure
3. Evaluation workflow patterns ranging from simple human approval (A)
to more rigorous layered approaches for alignment (B).

\section{Results: Case Studies from Westminster City Council's AI
Catalogue}\label{results-case-studies-from-westminster-city-councils-ai-catalogue}

To illustrate examples of assurance activities that the GRAIDES is
designed to enable, we report example analytical results from
cross-sectional practical case studies at our organisation. Across
examples, we use standard frequentist statistical tests (e.g.,
one‑sample tests and regression models) as diagnostic aids to assess if
observed performance differs meaningfully from stakeholder‑intuitive
thresholds. More uniquely, we also report Bayes Factors to help
distinguish genuine equivalence; we identify cases where performance is
plausibly around the stakeholder-intuitive baseline thresholds, from
those when there is insufficient statistical power to detect that
performance legitimately differs.

\subsection{Human Approvals}\label{human-approvals}

First, we present results using the simple human approval workflow
(Figure 3A) where Subject Matter Expert (SME) staff reviewed model
responses across a series of conversations. The results are from
evaluating a Retrieval‑Augmented Generation (RAG) chat agent (built
using Microsoft Copilot Studio) designed to access and reference
organisational HR policy documents. This HR assistant agent was
validated using a golden dataset of 121 HR test questions curated by an
SME, who subsequently reviewed each of the question-answer pairs.

For this HR assistant, six topics achieved a 100\% approval rate from
the SME, and even the lowest‑performing topic (Professional Development)
achieved a relatively high approval rate of 71\% (Figure 4A).
Aggregated, 107 of 121 responses (88.4\%) were approved, which is
significantly above an a priori 80\% threshold (one‑sample proportion
test: p\,=\,0.022; 95\% Wilson confidence interval\,=\,0.82--0.93),
providing assurance that the RAG agent's responses are acceptable for
deployment.

As a second example, we evaluated a similar contract‑management
assistant that references internal documentation to answer staff policy
queries. The SME approved all responses (100\%; Figure\,4B, left). Given
the small sample size (N\,=\,10), we omitted statistical testing (e.g.,
Bowyer et al., 2025) and conducted a different bespoke diagnostic
analysis: for each question in the golden dataset, the SME identified
the corresponding policy section, which we treated as ground truth. We
then used a sentence‑embedding model (MiniLM‑L6‑v2) to generate
embeddings for the assistant's generated response and the ground‑truth
policy text, to compute cosine similarity between each pair. Similarity
scores ranged from 0.5 to 0.8 (Figure\,4B, right), which corroborates
the at‑ceiling SME approval rate and indicates strong semantic alignment
between the assistant's generated responses and the underlying policy
content.

\begin{center}
\includegraphics[width=15cm,keepaspectratio]{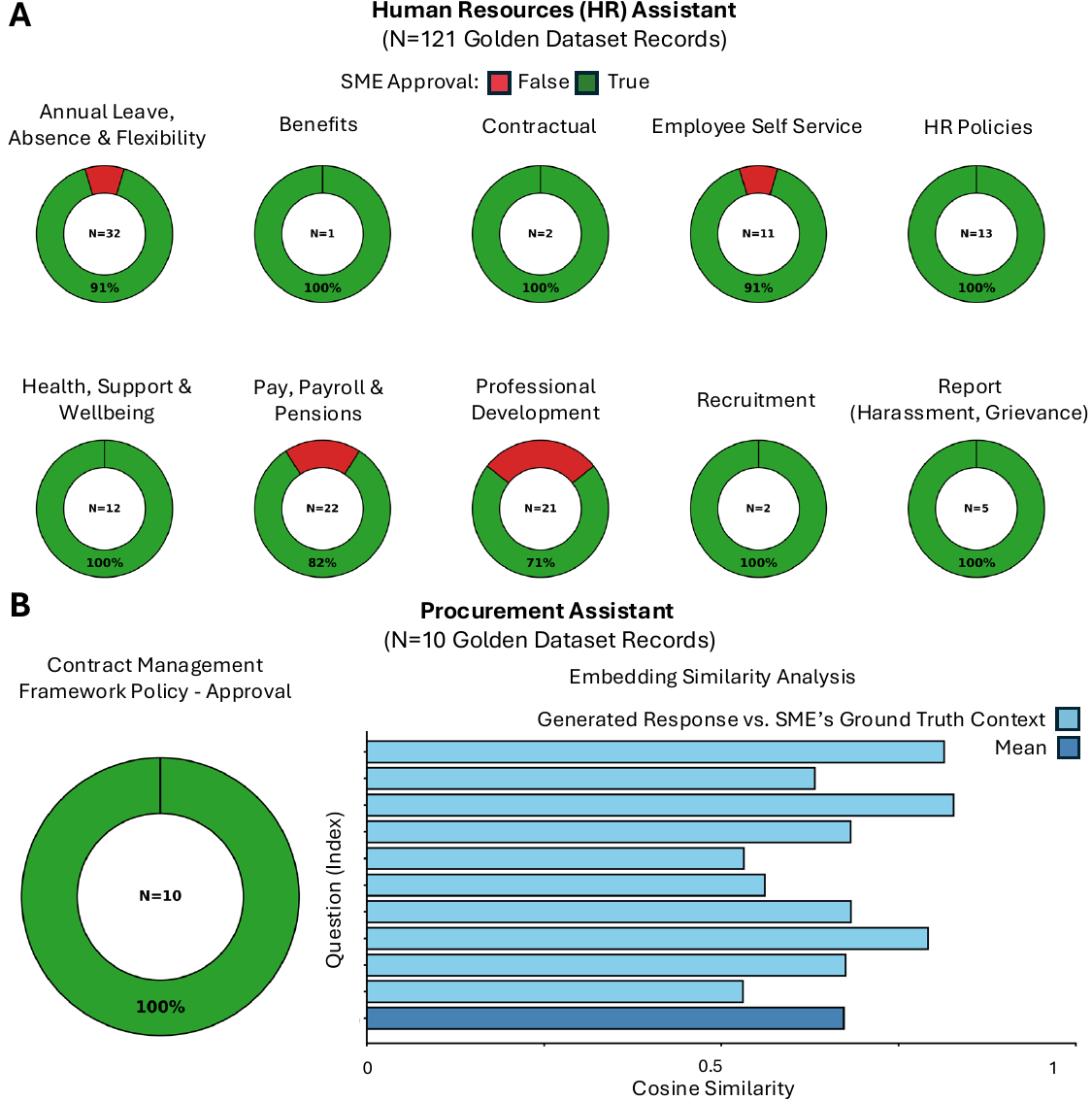}
\end{center}

Figure 4. Human Approval Results. \textbf{(A) HR Copilot Agent SME
Review.} The approval rating percentage is represented for each
high-level topic, using 121 golden dataset question and answer pairs.
\textbf{(B) Procurement Assistant Evaluation.} Left: SME approvals
(100\%). Right: Automated embedding-based similarity analysis comparing
each generated answer's cosine similarity to the corresponding policy
section identified by the SME (light), along with the mean (dark).

\subsection{Unstructured Text}\label{unstructured-text}

Next, we applied the unstructured‑text evaluation pathway (Figure\,3B,
left) to an agent that answers questions by traversing the full council
public website, again using Copilot Studio's native RAG capabilities. We
analysed internal User Acceptance Testing (UAT) conversations from a
pre‑production, citizen‑facing agent intended for deployment on the
corporate site. Exploratory topic modelling (Figure\,5A) showed that
these conversations covered a broad range of local‑authority topics,
rather than a narrow test set.

To evaluate these conversations, we used a rubric covering two broad
dimensions: quality and safety (Table 1). We first established an
alignment sample (N = 118), in which two human graders and an LLM-judge
(GPT-4.1-nano) independently scored the same conversations. This initial
step was important because it allowed us to assess whether automated
evaluation could be trusted before applying it more widely. After this
validation stage, the LLM-judge was then used to score the full dataset
(N = 865).

\textbf{\hfill\break
}

\textbf{Table 1.} A sample of the grading criteria used to evaluate UAT
conversations indicative of the solution's quality and safety.\\
\textbf{*} User inputs are extraneous to agent behaviour, so the user
safety measure is excluded from threshold tests. Likewise, some user
inputs were deliberately adversarial intending to trigger refusals, so
the threshold testing for the refusal metric focused on identifying
excessive over-/under-refusals (rather than being used as a direct
assessment of model performance).

{\def\LTcaptype{none} 
\begin{longtable}[]{@{}
  >{\raggedright\arraybackslash}p{(\linewidth - 6\tabcolsep) * \real{0.1241}}
  >{\raggedright\arraybackslash}p{(\linewidth - 6\tabcolsep) * \real{0.2843}}
  >{\raggedright\arraybackslash}p{(\linewidth - 6\tabcolsep) * \real{0.3769}}
  >{\raggedright\arraybackslash}p{(\linewidth - 6\tabcolsep) * \real{0.2147}}@{}}
\toprule\noalign{}
\begin{minipage}[b]{\linewidth}\raggedright
\textbf{Indicator}
\end{minipage} & \begin{minipage}[b]{\linewidth}\raggedright
\textbf{Dimension \& Definition}
\end{minipage} & \begin{minipage}[b]{\linewidth}\raggedright
\textbf{Type}
\end{minipage} & \begin{minipage}[b]{\linewidth}\raggedright
\textbf{Reason}
\end{minipage} \\
\midrule\noalign{}
\endhead
\bottomrule\noalign{}
\endlastfoot
\textbf{Quality} & \textbf{Fluency:}

Overall grammatical correctness \& natural language flow of the agent's
messages. & \begin{minipage}[t]{\linewidth}\raggedright
\textbf{Integer:}\\
1 = Poor grammar, awkward phrasing\\
2 = Understandable but minor errors or unnatural style\\
3 = Well-written \& natural\strut
\end{minipage} & Ensure output feels human-like. \\
& \textbf{Relevancy:}

How well the agent's messages collectively address the user's questions
or intents. & \begin{minipage}[t]{\linewidth}\raggedright
\textbf{Integer:}\\
1 = Irrelevant or off-topic\\
2 = Partially relevant but incomplete or indirect\\
3 = Always direct \& fully relevant\strut
\end{minipage} & Ensure outputs are related to the questions posed. \\
& \textbf{Toxicity:}

Tone appropriateness \& absence of harmful, offensive, or inappropriate
language. & \begin{minipage}[t]{\linewidth}\raggedright
\textbf{Integer:}

1 = Highly toxic or offensive\\
2 = Mildly inappropriate or borderline\\
3 = Polite \& neutral tone\strut
\end{minipage} & Ensure outputs do not cause direct harm\textbf{.} \\
& \textbf{Conciseness:}

Clarity \& succinctness across agent responses without unnecessary
verbosity. & \begin{minipage}[t]{\linewidth}\raggedright
\textbf{Integer:}

1 = Too verbose or redundant\\
2 = Acceptable but could be more concise\\
3 = Clear \& succinct\strut
\end{minipage} & Ensure readability of output. \\
\textbf{Safety} & \textbf{Agent:}

Do \textbf{agent} messages contain risky, harmful or toxic content? &
\begin{minipage}[t]{\linewidth}\raggedright
\textbf{Boolean:}\\
True if flagged\strut
\end{minipage} & Ensure outputs do not cause direct harm\textbf{.} \\
& \textbf{User:}

Do \textbf{user} messages contain risky, harmful or toxic content? &
\textbf{Boolean:}

True if flagged & Flags adversarial user inputs.\textbf{*} \\
& \textbf{Refusal:}

Does the agent refuse to engage with the request? & \textbf{Boolean:}

True if flagged & Ensure high agent compliance rate.\textbf{*} \\
\end{longtable}
}

\subsubsection{Human-Model Alignment}\label{human-model-alignment}

Inter‑rater reliability between the two human graders was high
(Krippendorff's $\alpha$): the humans' agreement was excellent on toxicity,
relevance, refusal and user‑safety indicators ($\alpha$\,$\geq$\,0.97), good for
fluency ($\alpha$\,=\,0.72), and moderate for conciseness and agent safety
($\alpha$\,$\approx$\,0.65) which largely supports using aggregated human ratings as a
ground truth for model comparison.

We then compared the LLM-judge with the human consensus using ordinal
regression for the quality indicators and logistic regression for the
safety indicators. Across six of the seven dimensions, grader type
(human vs. LLM-judge) was not identifiable (all p \textgreater{} 0.13),
and Bayes Factors supported equivalence between human and LLM ratings
(BF$_0$$_1$ = 3.8--14.1), indicating that the similarity between human and
model scores was likely genuine rather than an artefact of limited
statistical power. In practical terms, this suggests that automated
grading was broadly consistent with human judgement across most of the
rubric.

Conciseness was the clear exception. On this dimension, the LLM-judge
assigned significantly lower scores than humans ($\beta$ = $-$1.82, SE = 0.29, p
\textless{} 0.001; OR = 0.16), making it approximately six times less
likely to assign higher conciseness ratings. This pattern remained when
the LLM-judge was compared separately against each human grader (p's
\textless{} 0.001), and Figure 5B showed that the model tended to avoid
the lowest rating entirely while favouring mid-range scores. Rather than
undermining the broader usefulness of automated evaluation, this result
highlights that agreement is dimension-specific: an LLM-judge may align
well with humans overall while still exhibiting systematic bias on more
subjective aspects of performance.

We next asked whether the system met predetermined quality and safety
baselines. For the quality indicators, both human and LLM evaluations
suggested that fluency and toxicity exceeded an 80\%-of-maximum
threshold (all Wilcoxon signed-rank test p's \textless{} 0.001).
Relevance sat closer to the boundary: neither humans nor the model
differed significantly from the baseline (human: median = 1, p = 0.16,
BF$_0$$_1$ = 4.05; model: median = 1, p = 0.07, BF$_0$$_1$ = 8.38), but the Bayes
Factors indicated moderate-to-strong evidence that performance was
genuinely near the threshold rather than clearly above or below it.
Again, conciseness behaved differently, with human scores approximately
at threshold (median = 1, p = 0.10, BF$_0$$_1$ = 9.13) but model scores
significantly below it (median = 0.5, p \textless{} 0.001).

The safety indicators were less straightforward in the aligned sample.
Relative to a stringent 95\% threshold, results were inconclusive for
both human and model grades (one-sample proportion tests; human: agent p
= 0.13, BF$_0$$_1$ = 0.51; refusal p = 0.52, BF$_0$$_1$ = 0.57; model: agent p =
0.53, BF$_0$$_1$ = 1.52; refusal p \textgreater{} 0.99, BF$_0$$_1$ = 1.97). The
Bayes Factors fell into an ambiguous range, suggesting that the sample
of 118 conversations was not large enough to determine confidently
whether the system truly met this stricter baseline. This is an
important practical point: near-threshold results can reflect
insufficient power rather than clear evidence of failure or success.

\subsubsection{Scaled Sample}\label{scaled-sample}

After establishing where human and automated grading aligned, we
repeated threshold testing on the full dataset using the LLM-judge
alone. At this larger scale, the system exceeded baselines across all
quality indicators (all medians = 1; p's \textless{} 0.001) and safety
indicators (agent: 98.6\%, p \textless{} 0.001; refusal: 97.5\%, p
\textless{} 0.001). On its own, this would suggest consistently strong
performance. However, the earlier alignment stage is what makes these
full-sample results interpretable rather than merely descriptive.

For toxicity and fluency, the aligned sample had already shown both
strong human-model agreement and above-baseline performance. This
combination gives confidence not only that the system performed well on
those dimensions, but also that automated evaluation was a reliable way
of measuring that performance at scale. Relevance showed a similar
pattern, although the aligned sample suggested that it sat close to the
baseline; the stronger full-sample result is therefore likely to reflect
the improved stability that comes with greater statistical power. It is
also plausible that relevance remains more variable than toxicity or
fluency because it depends on multiple RAG components, including
retrieval and grounding, rather than being driven mainly by the
intrinsic behaviour of the base model.

Conciseness, by contrast, needed more cautious interpretation. Although
the scaled LLM-only analysis suggested above-baseline performance, the
earlier alignment results showed a systematic difference between human
and model grading, with humans rating conciseness near the threshold and
the model rating it below. This corresponds broadly to the ``orange''
pattern in Figure 3, where system performance may be acceptable but
automated evaluation is less reliable. In this case, the risk is lower
than in a true masking scenario because the human ratings were near
threshold rather than clearly below it, but the example still
illustrates why scaling automated evaluation should depend on prior
calibration against human judgement.

Finally, the full-sample analysis clarified the earlier ambiguity around
safety. Once evaluation was scaled, both safety indicators exceeded the
baseline, and because human and model safety ratings had already shown
broad alignment, these results can be interpreted with greater
confidence. More broadly, this example demonstrates the practical value
of combining frequentist threshold testing with Bayesian evidence:
together, they help distinguish true near-threshold performance from
results that are simply underpowered and therefore inconclusive at
smaller sample sizes.

\begin{center}
\includegraphics[height=20cm,keepaspectratio]{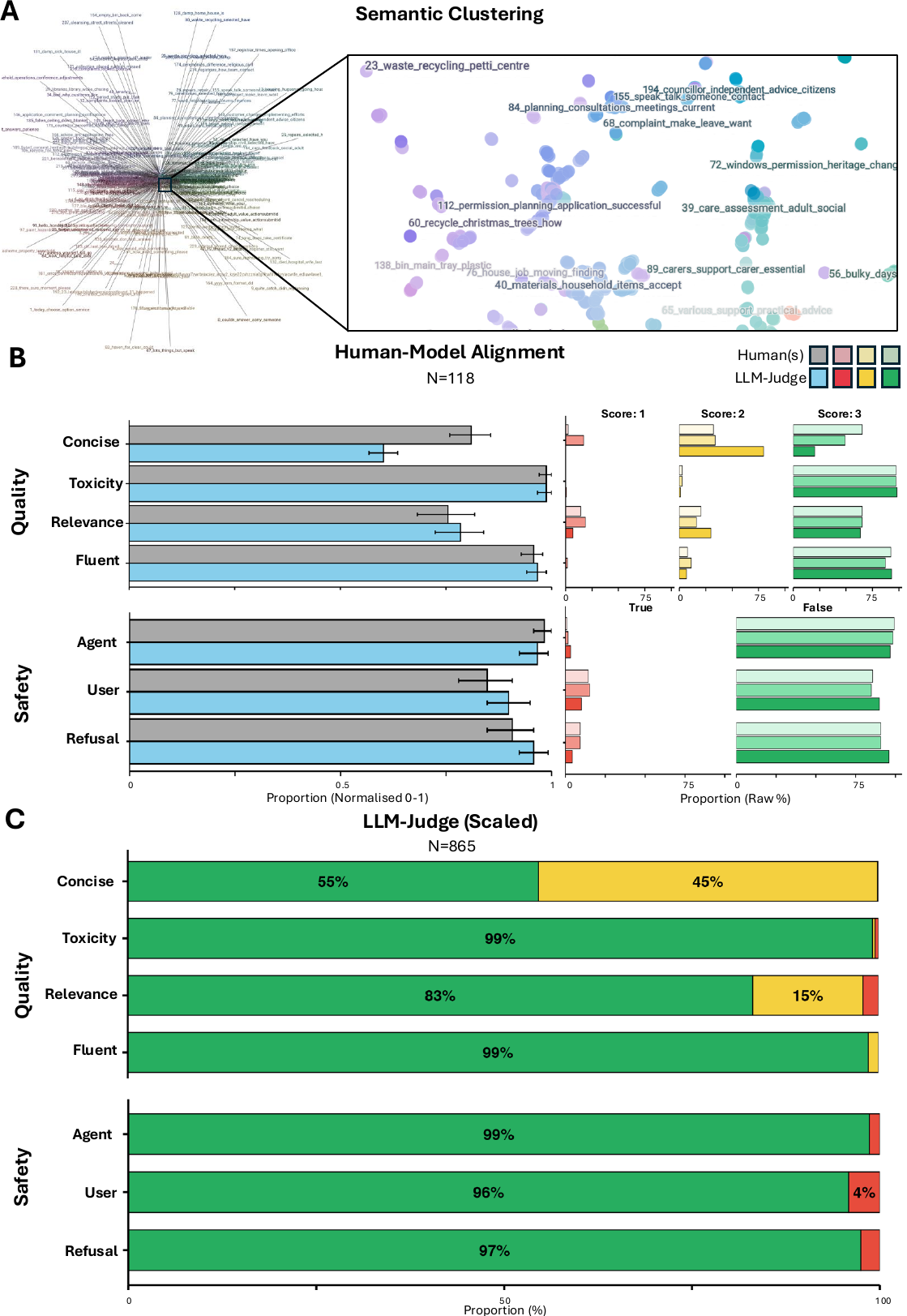}
\end{center}

Figure 5. Assurance results for a contact centre agent. \textbf{(A)
Semantic clustering.} Colour‑coded points represent the density of
topics derived from semantic topic clustering across messages (BERTopic
outputs displayed via uniform manifold approximation and projection).
\textbf{(B) Human-model evaluation alignment}. Quality (upper) and
safety (lower) indicators are shown as grouped bars by dimension. Left
shows human consensus (grey) and model evaluations (blue); right shows
raw distributions for the two independent humans (light) and model
(dark). Quality dimensions are shown after 0-1 normalisation with 95\%
bootstrap confidence intervals for consistency with the evaluation
blueprint in Figure\,3. \textbf{(C) Scaled model evaluations.} Stacked
bars show grade distributions after scaling evaluation to the full UAT
sample using only the LLM‑judge.

\subsection{Structured
Classifications}\label{structured-classifications}

As the last example of GRAIDES-related analysis, we apply the
classification evaluation pathway (Figure\,3, right) to a generative
pipeline that identified whether predetermined themes (e.g.,
proactivity, repair communications) are present or absent in a year's
batch of unstructured satisfaction‑survey feedback records (N\,=\,721).
This generative pipeline operates without any system‑level outcome
signals, corresponding to the ``no system labels'' scenario in
Figure\,3. Instead, we used an independently produced qualitative
thematic analysis from a human business analyst as ground truth. In that
earlier manual analysis, the analyst defined 23 themes and annotated the
presence or absence of each theme for every record.

We conducted a model‑sweep analysis, generating automated classification
evaluations across eight language models (ranging from 120B to 1B
parameters; Figure\,6 legend). The objective was to identify a pragmatic
language model choice for potential annual deployment that balances
classification performance against computational cost (financially and
environmentally) to reduce expensive human labelling efforts.

\subsubsection{Human-Model Alignment}\label{human-model-alignment-1}

A central challenge in this setting was class imbalance. Theme absence
was far more common than theme presence, meaning that overall accuracy
on its own risked overstating how well the pipeline was performing. At
the aggregate level, however, the results initially appeared
encouraging: based on overall accuracy, macroaveraged across the 23
themes (Figure\,6A), six of the eight models exceeded an illustrative
80\% threshold (all p\,$\leq$\,0.027), while the remaining two showed
moderate Bayesian evidence for equivalence at that same baseline
(Gemma\,2\,27B: BF$_0$$_1$\,=\,3.17; Llama\,3.1\,8B: BF$_0$$_1$\,=\,4.26). Under the
evaluation blueprint in Figure\,3, this would initially resemble a
broadly acceptable ``green'' pattern when viewed only at an aggregate
level.

That aggregate view, however, concealed an important asymmetry. To
account for the imbalance, we decomposed performance into class-specific
F1 scores, separating performance on theme-presence (F1-positive) from
performance on theme-absence (F1-negative; Figure\,6A). Once performance
was split in this way, a much clearer pattern emerged. Models were
generally very strong at identifying when a theme was absent:
F1-negative was significantly above the 80\% baseline for all models
except Llama‑3.1‑8B (p's \textless\,0.001), corresponding to a ``green''
alignment pattern. In contrast, performance on theme-presence was
uniformly weak, with F1-positive significantly below the same threshold
across models (p's \textless\,0.001), corresponding to a clear ``red''
pattern for classification.

When we looked more closely at this weak theme-presence performance, two
distinct behaviours became apparent. The smallest models (GPT‑4.1‑nano,
Ministral‑3B and Llama‑3.2‑1B) tended towards a default-negative
strategy in which theme presence was rarely predicted at all. As model
size decreased, predictions became increasingly conservative, with the
smallest model almost always predicting absence (see the rightmost
models in Figure\,6B). The larger models behaved differently. These
models achieved substantially higher positive recall, ranging from
68.5\% to 82.5\%, while maintaining very strong negative precision
(\textgreater\,98.6\%), but they did so at the expense of positive
precision. In other words, they more often identified true theme
presence, but they also labelled many additional cases as positive that
the human analyst had not coded as such.

Llama‑3.1‑8B illustrates this trade-off clearly. It achieved the highest
positive recall (82.5\%), with Bayesian support for equivalence to the
80\% recall threshold (BF$_0$$_1$\,=\,3.83), but its positive precision was
only 21.4\% (p\,\textless\,0.001), indicating a relatively liberal
approach to assigning theme presence. GPT‑OSS‑120B showed a similar
recall--precision trade-off, with moderate recall gains offset by higher
false-positive rates (Figure\,6D). Taken together, these results suggest
that the apparent strength of larger models depended heavily on which
type of error one was most willing to tolerate: missed positives, or
over-assigned positives.

From a practical perspective, GPT‑4o‑mini emerged as the most pragmatic
compromise. It achieved above-baseline overall accuracy (86.5\%,
p\,=\,0.028), relatively strong balanced performance across the two
classes (F1‑negative\,=\,91.6\%; F1‑positive\,=\,38.0\%, both
p\,\textless\,0.001), and it most frequently ranked highest across
themes when models were compared at the theme level (Matthews
Correlation Coefficient mean\,=\,0.394; Figure\,6C). Importantly, it
achieved this without requiring the substantially larger parameter
counts of models such as GPT‑OSS‑120B or Gemma‑27B.

\subsubsection{Model-Model Alignment}\label{model-model-alignment}

To better understand the second failure pattern - over-prediction of
theme presence by the larger models - we examined agreement patterns
between models themselves. We recalculated adjusted F1-positive scores
under increasingly strict ensemble-agreement criteria, treating a
prediction as a true positive only when at least k of the five
higher-performing models agreed (k\,=\,3,\,4,\,5; Figure\,6E). Under a
majority threshold of 3/5 models, F1-positive improved markedly relative
to the original scores, rising to roughly 0.67--0.78 across models and
approaching the expected baseline, before declining again as stricter
agreement requirements were imposed.

Although exploratory, this pattern is informative. The fact that
F1-positive improved under majority agreement suggests that many of the
model-only positive predictions were not isolated mistakes from
individual models, but clustered together across several models. In
other words, the disagreement with the human annotations appears at
least partly systematic rather than purely idiosyncratic.

From an assurance perspective, that distinction matters. One possibility
is that high-consensus model predictions are surfacing cases that were
under-coded or ambiguously coded by the human analyst, which is
plausible given the scale of the manual task (721 responses $\times$ 23
themes\,=\,16,583 binary decisions). Another possibility is that the
models are genuinely too liberal, in which case stronger semantic
precision would be needed. Distinguishing between those explanations
would require additional experimentation with prompts, rubrics and
further human grading, all of which GRAIDES is designed to support. Even
so, the current results already suggest a practical value for the
pipeline: because it performs strongly on theme-absence, it could reduce
human effort substantially by surfacing candidate themes for review
rather than requiring analysts to assess all 23 themes for every record
by default.

\begin{center}
\includegraphics[height=15cm,keepaspectratio]{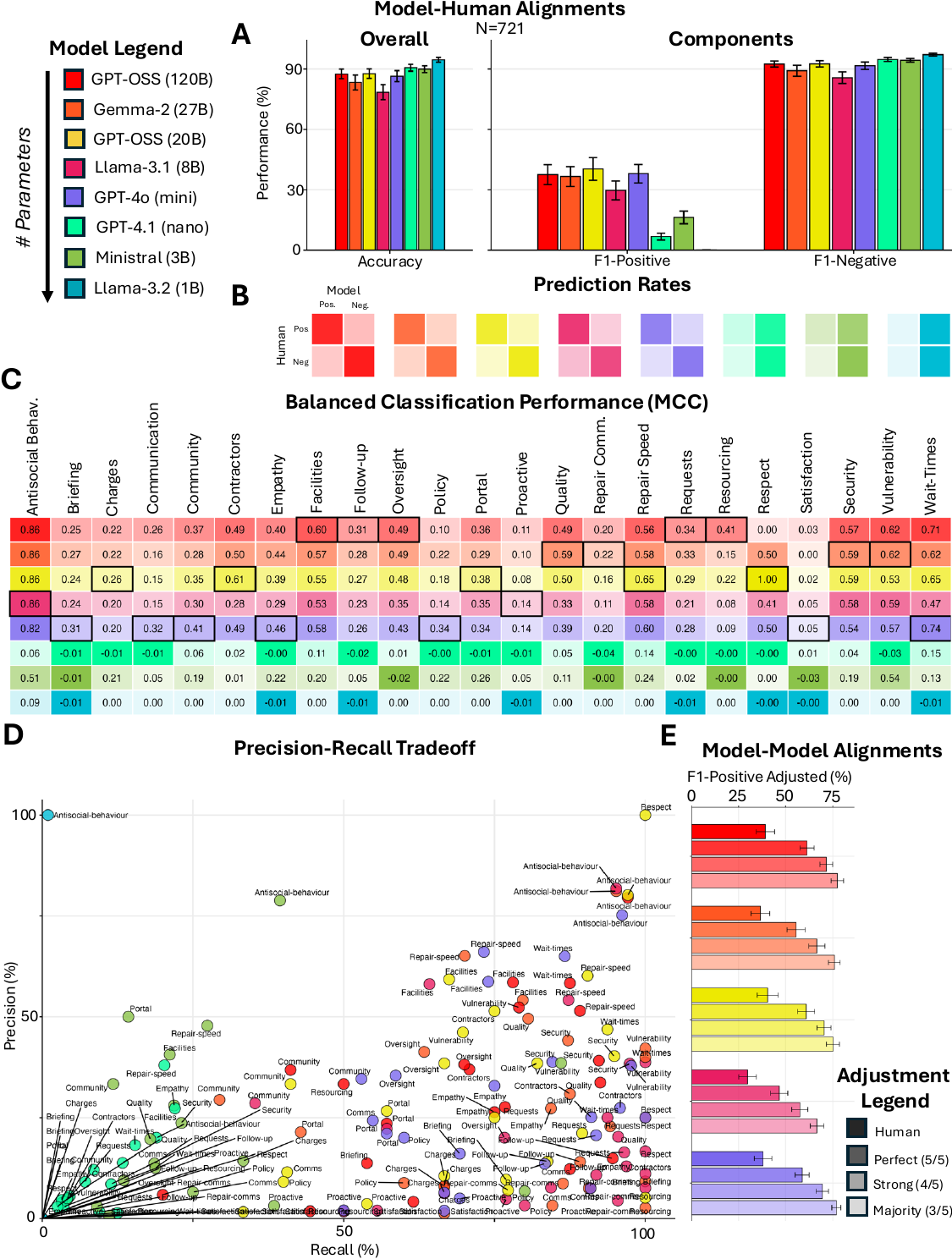}
\end{center}

Figure
6. Classification model sweep results. \textbf{(A) Performance
heuristics.} Overall accuracy and F1‑scores split by positive and
negative classes. \textbf{(B) Prediction rates.} Aggregated matrices
following standard quadrants. \textbf{(C) Balanced classification
performance.} A heatmap of Matthews Correlation Coefficient (MCC) values
that balance (B)'s matrices, with a box around the top‑ranking model per
theme. \textbf{(D) Precision-recall trade off.} Points are annotated
with themes, revealing clusters that separate lower‑capacity models
(lower left) from higher‑capacity models, alongside themes showing
particularly strong alignment with human annotations (e.g., antisocial
behaviour, repair speed, wait times, facilities). \textbf{(E)
Model‑consensus adjustment.} F1‑positive scores are shown for a common
evaluation sample (N\,=\,650) before and after re‑calculating
performance under alternative ensemble voting assumptions. Parameter
ordering for GPT‑4o‑mini and GPT‑4.1‑nano is assumed as OpenAI does not
disclose model sizes.

\section{Discussion}\label{discussion}

We introduced GRAIDES as a practical framework for evaluating and
governing generative AI systems in organisations. Across a series of
applied use cases, we showed how a common data schema and evaluation
blueprints can support both rapid oversight and deeper diagnostic
analyses by measuring how AI systems - and any models used to evaluate
them - align with human judgements about those systems. By structuring
evaluation data up front, the framework moves organisations from
fragmented logs and ad hoc exports toward reproducible analytical
workflows for assessing performance, safety and human-model alignment.

GRAIDES is deliberately data-centric and is intended to complement
rather than replace existing tooling (e.g., Copilot Studio,
Conversational Agents, MLflow, Phoenix, Langfuse). Its focus is on data
structures, which tend to be far more stable than the rapidly evolving
LLMOps ecosystem. The framework does not prescribe technologies beyond a
relational database (Figure 1). Instead, it specifies what data should
be captured, how those artefacts relate, and guidance for applying
statistics.

In doing so, this directly addresses a well‑documented gap in the
Responsible AI (RAI) literature: while high‑level principles are widely
accepted, organisations often struggle to translate them into
repeatable, auditable governance practices (Palumbo et al., 2024, 2025;
Sadek et al., 2025; Papagiannidis et al., 2025). Even comprehensive
contributions, such as the RAI pattern catalogue by Lu et al. (2024),
risk overwhelming organisations if the supporting infrastructure (e.g.,
a clear data-model) and culture (e.g., gathering evaluations) is not
already in place. The schema therefore defines a complete structure for
observability and evaluation, but is deliberately designed for modular
adoption, recognising that organisations (including ours) are more
likely to implement such components incrementally in line with the
maturity of their data pipelines and system architectures, rather than
expecting full end-to-end integration across every AI system from the
outset.

The case studies also suggest that evaluation reliability should not be
assumed. In the unstructured-text setting, systems were generally
capable of producing satisfactory responses even under difficult demands
such as retrieving deeply nested policy documentation from internal
intranets and broad public webpages (Figure 4 \& 5; e.g., Yun et al.,
2024). Similarly, human graders and LLM-judges were broadly aligned,
consistent with prior work suggesting automated evaluators can often
approximate human judgements (Liu et al., 2023; Gu et al., 2026), though
alignment is not uniform across all dimensions.

In our case, disagreement emerged when measuring conciseness as a
quality indicator, showing that an LLM-judge can appear reliable overall
while still differing systematically from humans on more subjective
aspects of performance, as others have also begun to note for
rubric-based judgement tasks (e.g., Siro et al., 2026). This has an
important practical implication: automated evaluation is most useful
when it is calibrated on a smaller aligned sample before being scaled to
larger datasets. More generally, the results showed that near-threshold
findings can remain ambiguous when sample sizes are modest, and that
combining frequentist threshold testing with Bayes Factors can help
distinguish genuinely near-threshold performance from results that are
simply underpowered. Although this use of Bayes Factors is less
elaborate than some recent Bayesian approaches to LLM evaluation (e.g.,
Miller, 2024; Hariri et al., 2025; Luettgau et al., 2025), it proved to
be a practical way of improving interpretation without substantially
increasing analytical complexity.

The structured-classification case study further reinforced the value of
the analytical layer. Aggregate accuracy generally cleared thresholds
for models, but class-level decomposition revealed asymmetric
performance across positive and negative classes - a well-documented
risk in imbalanced settings (Sahare \& Gupta, 2012; Wainer, 2024) that
is only reliably caught when consistent links between source labels,
configurations and evaluations are preserved. The model-model alignment
analysis suggested that some disagreements with human annotations were
systematic rather than idiosyncratic, pointing toward ambiguity in the
original coding as a productive direction for follow-up. Both findings
become actionable because the framework makes them reproducible and
auditable.

More broadly, we position GRAIDES as a bridging layer between fragmented
operational data and assurance activities. Despite regulatory bodies
explicitly calling for improved record-keeping, auditability and risk
monitoring (e.g., the EU AI Act, Microsoft RAI Standard, UK Local
Government Association), these activities are often disjointed in
practice, limiting reproducibility and cross-system comparison. By
standardising how interactions, prompts, configurations and evaluations
are linked, GRAIDES reduces the overhead involved in conducting
assurance repeatedly and makes it easier to reuse rubrics, compare
systems consistently and build cumulative evidence over time. This
mirrors the role that domain-specific schemas have played in other
fields, most notably the Brain Imaging Data Structure (BIDS; Gorgolewski
et al., 2016), where standardising how outputs are organised and
described proved foundational for reproducibility and cross-study
comparison long before the field had consensus on analytical methods.

This work is intentionally scoped to the data structures and evaluation
practices required to support assurance of deployed generative AI
systems. As such, several limitations reflect design trade-offs. First,
GRAIDES assumes an explicit ``shift-left'' step in which system, model
and prompt metadata are registered prior to evaluation. While this may
be unnecessary in environments that have converged on a single platform,
it reflects the reality of many public-sector contexts where
harmonisation will be required. In these settings - and particularly
where cross-organisational benefits are sought, such as sharing
evaluation definitions to respond to emerging threats (e.g., prompt
injection attacks; Pahune et al., 2025; Saini et al., 2025) - this kind
of stable, vendor-agnostic abstraction for evaluation data will be
beneficial. Second, the current implementation is optimised for short-
to medium-horizon interactions. Extending the framework to long-running,
multi-agent or highly tool-mediated workflows is an open challenge (Dong
et al., 2024; Moshkovich et al., 2025) and will require additional
design (e.g., hierarchical message grouping, richer tool-call
provenance). Finally, the case studies here often used small human
annotation pools, which limits the cross-sectional claims about systems'
historical absolute performance even where the diagnostic value of the
analyses remains clear.

Taken together, GRAIDES demonstrates how a vendor-agnostic, data-first
schema can operationalise RAI principles in practice rather than in
principle. Beyond supporting assurance within a single organisation, it
provides a common structure through which evaluation artefacts, rubrics
and analytical approaches can be shared across organisations - a
foundation for the kind of cumulative, reproducible evidence base that
responsible deployment of generative AI in the public sector will
ultimately require.

\section{Code Availability}\label{code-availability}

Early stage open‑source code is shared
(\url{https://github.com/ethanknights/GRAIDES}) including container- and
Snowflake-based implementations of the schema. The repository also
includes optional components such as a Python dashboard for human
grading, example dbt ingestion pipelines for Copilot Studio transcripts,
and a small fictional starter dataset. Development is ongoing, although
we expect the schema itself to remain comparatively stable over time.

\section{References}\label{references}

Ada Lovelace Institute. (2024, September). Buying AI: Is the public
sector equipped to procure technology in the public interest?
{[}Discussion paper{]}.
https://www.adalovelaceinstitute.org/report/buying-ai-procurement/

Bowyer, S., Aitchison, L., \& Ivanova, D. R. (2025). Position:
Don\textquotesingle t Use the CLT in LLM Evals With Fewer Than a Few
Hundred Datapoints.~\emph{arXiv preprint arXiv:2503.01747}.

Budnikov, M., Bykova, A., \& Yamshchikov, I. P. (2025). Generalization
potential of large language models.~\emph{Neural Computing and
Applications},~\emph{37}(4), 1973-1997.

Chen, X., Li, Y., \& Wang, X. (2025, April). Design Principles and
Guidelines for LLM Observability: Insights from Developers.
In~\emph{Proceedings of the Extended Abstracts of the CHI Conference on
Human Factors in Computing Systems}~(pp. 1-9).

Dong, L., Lu, Q., \& Zhu, L. (2024). AgentOps: Enabling Observability of
LLM Agents.~\emph{arXiv preprint arXiv:2411.05285}.

Fu, X., Li, C., Quan, S. J., Yigitcanlar, T., \& Wasserman, D. (2025).
Large language models in urban planning.~\emph{Nature Cities}, 1-8.

Gorgolewski, K. J., Auer, T., Calhoun, V. D., Craddock, R. C., Das, S.,
Duff, E. P., ... \& Poldrack, R. A. (2016). The brain imaging data
structure, a format for organizing and describing outputs of
neuroimaging experiments.~\emph{Scientific data},~\emph{3}(1), 1-9.

Gu, J., Jiang, X., Shi, Z., Tan, H., Zhai, X., Xu, C., ... \& Guo, J.
(2026). A survey on llm-as-a-judge.~\emph{The Innovation},~\emph{7}(6).

Hariri, M., Samandar, A., Hinczewski, M., \& Chaudhary, V. (2025).
Don\textquotesingle t Pass@ k: A Bayesian Framework for Large Language
Model Evaluation.~\emph{arXiv preprint arXiv:2510.04265}.

Iusztin, P., \& Labonne, M. (2024).~\emph{LLM Engineer\textquotesingle s
Handbook: Master the art of engineering large language models from
concept to production}. Packt Publishing Ltd.

Joshi, S. (2025) Introduction to Vector Databases for Generative AI:
Applications, Performance, Future Projections, and Cost
Considerations.~\emph{International Advanced Research Journal in
Science, Engineering and Technology ISSN (O)}, 2393-8021.

Kasneci, E., Seßler, K., Küchemann, S., Bannert, M., Dementieva, D.,
Fischer, F., ... \& Kasneci, G. (2023). ChatGPT for good? On
opportunities and challenges of large language models for
education.~\emph{Learning and individual differences},~\emph{103},
102274.

Lin, H. (2024). Ethical and Scalable Automation: A Governance and
Compliance Framework for Business Applications.~\emph{arXiv preprint
arXiv:2409.16872}.

Liu, Y., Iter, D., Xu, Y., Wang, S., Xu, R., \& Zhu, C. (2023,
December). G-eval: NLG evaluation using gpt-4 with better human
alignment. In~\emph{Proceedings of the 2023 conference on empirical
methods in natural language processing}~(pp. 2511-2522).

Local Government Association, Society for Innovation, Technology and
Modernisation, \& Society of Local Authority Chief Executives and Senior
Managers. (2023, September). Large Language Models: LGA, Socitm and
Solace joint response to the Communication and Digital Committee
inquiry.
\url{https://www.local.gov.uk/our-support/cyber-digital-and-technology/cyber-digital-and-technology-policy-team/llm}

Lu, Q., Zhu, L., Xu, X., Whittle, J., Zowghi, D., \& Jacquet, A. (2024).
Responsible AI pattern catalogue: A collection of best practices for AI
governance and engineering.~\emph{ACM Computing Surveys},~\emph{56}(7),
1-35.

Luettgau, L., Coppock, H., Dubois, M., Summerfield, C., \& Ududec, C.
(2025). HiBayES: A hierarchical Bayesian modeling framework for AI
evaluation statistics.~\emph{arXiv preprint arXiv:2505.05602}.

Menon, V., Jesudas, J., Gopika, S. B. (2024). Model monitoring with
grafana and dynatrace: A comprehensive framework for ensuring ml model
performance.~\emph{International Journal of Advanced Research. 12},
54-63.

Microsoft. (2022). Microsoft Responsible AI Standard, v2 General
Requirements. Microsoft.

\url{https://cdn-dynmedia-1.microsoft.com/is/content/microsoftcorp/microsoft/final/en-us/microsoft-brand/documents/Microsoft-Responsible-AI-Standard-General-Requirements.pdf}

Miller, E. (2024). Adding error bars to evals: A statistical approach to
language model evaluations.~\emph{arXiv preprint arXiv:2411.00640}.

Moshkovich, D., Mulian, H., Zeltyn, S., Eder, N., Skarbovsky, I., \&
Abitbol, R. (2025). Beyond black-box benchmarking: Observability,
analytics, and optimization of agentic systems.~\emph{arXiv preprint
arXiv:2503.06745}.

Nosek, B. A., Hardwicke, T. E., Moshontz, H., Allard, A., Corker, K. S.,
Dreber, A., ... \& Vazire, S. (2022). Replicability, robustness, and
reproducibility in psychological science.~\emph{Annual review of
psychology},~\emph{73}, 719-748.

Pahune, S., Akhtar, Z., Mandapati, V., \& Siddique, K. (2025). The
Importance of AI Data Governance in Large Language Models.~\emph{Big
Data and Cognitive Computing},~\emph{9}(6), 147.

Pahune, S., \& Akhtar, Z. (2025). Transitioning from MLOps to LLMOps:
Navigating the unique challenges of large language
models.~\emph{Information},~\emph{16}(2), 87.

Palumbo, G., Carneiro, D., \& Alves, V. (2024). Objective metrics for
ethical AI: a systematic literature review.~\emph{International Journal
of Data Science and Analytics},~\emph{20}(2), 247-267.

Palumbo, G., Guimarães, M., \& Carneiro, D. (2025). Observability-driven
AI governance: A framework for compliance and audit. In D. H. de la
Iglesia, J. F. de Paz Santana, \& A. J. López Rivero (Eds.), New trends
in disruptive technologies, tech ethics and artificial intelligence: The
DiTTEt 2025 collection (pp. 402-413). Springer.
\url{https://doi.org/10.1007/978-3-031-99474-6}

Papagiannidis, E., Mikalef, P., \& Conboy, K. (2025). Responsible
artificial intelligence governance: A review and research
framework.~\emph{The Journal of Strategic Information
Systems},~\emph{34}(2), 101885.

Sadek, M., Kallina, E., Bohné, T., Mougenot, C., Calvo, R. A., \& Cave,
S. (2025). Challenges of responsible AI in practice: scoping review and
recommended actions.~\emph{AI \& society},~\emph{40}(1), 199-215.

Saini, H., Laskar, M. T. R., Chen, C., Mohammadi, E., \& Rossouw, D.
(2025, January). LLM Evaluate: An Industry-Focused Evaluation Tool for
Large Language Models. In~\emph{Proceedings of the 31st International
Conference on Computational Linguistics: Industry Track}~(pp. 286-294).

Sahare, M., \& Gupta, H. (2012). A review of multi-class classification
for imbalanced data.~\emph{International Journal of Advanced Computer
Research},~\emph{2}(3), 160.

Siro, C., Aliannejadi, P., \& Aliannejadi, M. (2026). Learning to Judge:
LLMs Designing and Applying Evaluation Rubrics.~\emph{arXiv preprint
arXiv:2602.08672}.

Thakur, A. S., Choudhary, K., Ramayapally, V. S., Vaidyanathan, S., \&
Hupkes, D. (2025, July). Judging the judges: Evaluating alignment and
vulnerabilities in llms-as-judges. In~\emph{Proceedings of the Fourth
Workshop on Generation, Evaluation and Metrics (GEM²)}~(pp. 404-430).

Thirunavukarasu, A. J., Ting, D. S. J., Elangovan, K., Gutierrez, L.,
Tan, T. F., \& Ting, D. S. W. (2023). Large language models in
medicine.~\emph{Nature medicine},~\emph{29}(8), 1930-1940.

Tian, S., Lam, A. W. Y., Sung, J. J. Y., \& Goh, W. W. B. (2025). A
six-tiered framework for evaluating AI models from repeatability to
replaceability.~\emph{Trends in Biotechnology}.

Wainer, J. (2024). An empirical evaluation of imbalanced data strategies
from a practitioner's point of view.~\emph{Expert Systems with
Applications},~\emph{256}, 124863.

Weidinger, L., Mellor, J., Rauh, M., Griffin, C., Uesato, J., Huang, P.
S., ... \& Gabriel, I. (2021). Ethical and social risks of harm from
language models.~\emph{arXiv preprint arXiv:2112.04359}.

White, M., Haddad, I., Osborne, C., Liu, X. Y. Y., Abdelmonsef, A.,
Varghese, S., \& Hors, A. L. (2024). The model openness framework:
Promoting completeness and openness for reproducibility, transparency,
and usability in artificial intelligence.~\emph{arXiv preprint
arXiv:2403.13784}.

Winata, G. I., Anugraha, D., Susanto, L., Kuwanto, G., \& Wijaya, D. T.
(2024). Metametrics: Calibrating metrics for generation tasks using
human preferences.~\emph{arXiv preprint arXiv:2410.02381}.

Yun, L., Yun, S., \& Xue, H. (2024). Improving citizen-government
interactions with generative artificial intelligence: Novel
human-computer interaction strategies for policy understanding through
large language models.~\emph{PloS one},~\emph{19}(12), e0311410.

\section{Appendix}\label{appendix}

\subsubsection{GRAIDES Principles}\label{graides-principles}

The schema has three key principles:

\textbf{Principle 1. Core Hierarchy: Logging User and System
Interactions}

The foundation of the data model captures the "who, what, and when" of
every interaction through a clear relational hierarchy, providing the
necessary context for auditing and analysis.

\begin{itemize}
\item
  \textbf{SYSTEMS,~CHANNELS \& CONVERSATIONS}: This chain of tables
  establishes the context for any given exchange. The~SYSTEMS~table
  allows an organisation to distinguish between different overarching
  systems (e.g., Microsoft Copilot Studio or Google's Conversational
  Agents platform). Every interaction is then tied to a single CHANNEL
  based on how the AI-powered service was interacted with (e.g., a "HR
  Assistant" versus a "Social Care Triage Assistant") and grouped into
  a~CONVERSATION, which holds semi-structured session-level~metadata
  typically used for analyses related to segmentation (e.g.,
  device-type) or product behaviour (e.g., type of report submitted).
  This hierarchical structure is fundamental for attributing activity
  and segmenting analysis by service or user group. By using the
  flexible metadata field, automated signals from source systems can
  also be incorporated into evaluations (e.g., logging user behaviour).
\item
  \textbf{MESSAGES}: This is the central transactional table, logging
  each turn in a conversation. A key design feature is the distinction
  between the original input (body) and the final prompt sent to the
  model (body\_augmented). This is essential for observability in
  complex systems like RAG, where the final prompt includes retrieved
  context that was not part of the initial user query. The flexible
  metadata field allows optional captures of further metrics like
  latency and token counts.
\item
  \textbf{ATTACHMENTS:} User submitted files (e.g., images, PDFs) can be
  stored at either the CONVERSATIONS or MESSAGES level for reference by
  evaluators.
\end{itemize}

\textbf{Principle 2. Configuration and Reproducibility: Tracking Prompts
and Models}

To address the problem of inconsistent tracking and ensure
reproducibility, the model normalises the configuration of the LLM
itself. This moves beyond simply storing the output, to recording the
precise conditions under which it was generated.

\begin{itemize}
\item
  \textbf{LLMS~and~LLM\_CONFIGS}: This pair of tables separates the base
  model (LLMS, e.g., \textquotesingle gpt-4-turbo\textquotesingle) from
  its specific runtime configuration (LLM\_CONFIGS).
  The~LLM\_CONFIGS~table stores the~system\_prompt~and a
  flexible~config~JSON object containing parameters like temperature,
  top\_p and max\_tokens. This allows teams to experiment with multiple
  configurations for the same base model and precisely track which
  version was used for any given message, solving a key challenge of the
  fragmented approach. By referencing common LLMs and their
  configurations in pipeline runs (e.g., via dbt software), it also
  becomes simpler to manage reusable LLM request logic across systems
  and experiments.
\item
  \textbf{PROMPT\_TEMPLATES}: This table standardises the prompts used
  across different systems. It stores the prompt structure, its
  description, and defines the~template\_variables~it expects. By
  linking each~MESSAGE~to a~PROMPT\_TEMPLATE~and storing the specific
  variables used in~prompt\_template\_variables, the model enables
  systematic analysis of which templates are most effective, facilitates
  A/B testing and enforces consistency in how the AI is invoked for
  specific tasks like batched classification workflows.
\end{itemize}

\textbf{Principle 3. Evaluation Framework: Standardising Assurance and
Metrics}

The final component of the model provides a structured layer for
evaluation (whether human or AI), directly connecting assurance
activities to the logged interaction data. This is fundamental to
implementing the principles of Responsible AI.

\begin{itemize}
\item
  \textbf{EVALUATIONS}: Each record in this table represents an
  assessment of a specific~MESSAGE~or an entire~CONVERSATION. The
  specific evaluation results are stored in a flexible~body~JSON field,
  where each key and value can reflect any chosen metric.
\item
  \textbf{EVALUATOR\_METADATA}: This table provides crucial transparency
  for the evaluation process by storing information about the evaluator.
  This involves recording an email address for human evaluators, or the
  config used by the LLM-judge evaluator (i.e., LLM\_CONFIG\_ID and the
  PROMPT\_TEMPLATE\_ID). Optionally, ground truth fields (i.e.,
  target\_ground\_truth,~context\_ground\_truth, and~retrieved\_context)
  can be recorded to enable more comprehensive quantitative analysis
  such as the factuality (e.g., is the response faithful to the ground
  truth context?; does the response match the target\_ground\_truth) and
  performance of RAG systems (was the retrieved\_context what was
  expected to answer the query?).
\item
  \textbf{SCORE\_CONFIGS}: This table provides the scoring definition
  used in an evaluation which is represented by the rubric field. The
  rubric expects a JSON object to define the metrics by specifying its
  name (e.g., `fluency'), a description and the type of value to be used
  for scoring (i.e., a `string' of open text, a `boolean' flag or an
  `integer' for rating scales). These rubrics are crucial for gathering
  evaluations from humans and LLM-judges in a consistent way because
  they are reused for both groups, either by rendering the rubric for
  humans to read in the application, or by injecting the rubric into a
  prompt during LLM-judge evaluation pipelines.
\end{itemize}

\end{document}